\documentclass{article}

\usepackage[final]{neurips_2025_ml4ps}
\PassOptionsToPackage{numbers,sort&compress}{natbib}

\usepackage[utf8]{inputenc} % allow utf-8 input
\usepackage[T1]{fontenc}    % use 8-bit T1 fonts
\usepackage{hyperref}       % hyperlinks
\usepackage{url}            % simple URL typesetting
\usepackage{booktabs}       % professional-quality tables
\usepackage{amsfonts}       % blackboard math symbols
\usepackage{nicefrac}       % compact symbols for 1/2, etc.
\usepackage{microtype}      % microtypography
\usepackage{xcolor}         % colors
\usepackage{amsmath}
\usepackage{natbib}
\usepackage{graphicx}
\usepackage{caption}
\usepackage{subcaption}
\usepackage{multirow}
\usepackage{aas_macros}

\DeclareMathOperator{\E}{\mathbb{E}}
\RequirePackage{algorithm}
\RequirePackage{algorithmic}

\title{
    Uncovering Physical Drivers of Dark Matter Halo Structures with Auxiliary‑Variable‑Guided Generative Models
}

\author{%
  Arkaprabha Ganguli \thanks{equal contribution}\\
  Mathematics and Computer Science Division\\
  Argonne National Laboratory\\
  Lemont, IL 60439 \\
  \texttt{aganguli@anl.gov} 
  \And
  Anirban Samaddar $^*$\\
  Mathematics and Computer Science Division\\
  Argonne National Laboratory\\
  Lemont, IL 60439 \\
  \texttt{asamaddar@anl.gov} 
  \And
  Florian K\'eruzor\'e \\
  High Energy Physics Division\\ Argonne National Laboratory \\
  Lemont, IL 60439, USA \\
  \texttt{fkeruzore@anl.gov} \\
   \And
  Nesar Ramachandra\\
  Computational Science Division\\
  Argonne National Laboratory\\
  Lemont, IL 60439 \\
  \texttt{nramachandra@anl.gov} 
  \And
  Julie Bessac \\
  Computational Science Center\\ Data, Analysis and Visualization Group\\
  National Renewable Energy Laboratory\\
  Golden, CO, USA \\
  \texttt{julie.bessac@nrel.gov} 
  \And
  Sandeep Madireddy \\
  Mathematics and Computer Science Division\\
  Argonne National Laboratory\\
  Lemont, IL 60439 \\
  \texttt{smadireddy@anl.gov} 
  \And
  Emil Constantinescu \\
  Mathematics and Computer Science Division\\
  Argonne National Laboratory\\
  Lemont, IL 60439 \\
  \texttt{emconsta@anl.gov} 
}

\begin{document}

\maketitle

% \title{
%     % Uncovering Physical Drivers of Cosmic Microwave Background Halo Structures with Auxiliary‑Variable‑Guided Variational Autoencoders
%     Uncovering Physical Drivers of Dark Matter Halo Structures from maps of the thermal Sunyaev-Zel'dovich effect with Auxiliary‑Variable‑Guided Variational Autoencoders
% }
\begin{abstract}
Deep generative models (DGMs) compress high-dimensional data but often entangle distinct physical factors in their latent spaces. We present an auxiliary-variable-guided framework for disentangling representations of thermal Sunyaev–Zel’dovich (tSZ) maps of dark matter halos. 
We introduce halo mass and concentration as auxiliary variables and apply a lightweight alignment penalty to encourage latent dimensions to reflect these physical quantities. 
% To overcome the limitations of VAEs in 
To generate sharp and realistic samples, we 
% extend this approach using 
extend latent conditional flow matching (LCFM), a state-of-the-art generative model, to enforce disentanglement in the latent space. 
% The resulting hybrid 
Our \textbf{D}isentangled \textbf{L}atent-\textbf{CFM} (\texttt{DL-CFM}) model recovers the established mass-concentration scaling relation and identifies latent space outliers that may correspond to unusual halo formation histories. By linking latent coordinates to interpretable astrophysical properties, our method transforms the latent space into a diagnostic tool for cosmological structure. This work demonstrates that auxiliary guidance preserves generative flexibility while yielding physically meaningful, disentangled embeddings, providing a generalizable pathway for uncovering independent factors in complex astronomical datasets.
\end{abstract}

\section{Introduction}
Deep generative models (DGMs)—including variational autoencoders (VAEs), normalizing flows, and diffusion models are indispensable for modeling complex, high-dimensional scientific data. However, when applied to scientific datasets, heterogeneous in modality, fidelity, and accuracy, with stochastic measurements and multi-scale structure, DGMs often lack interpretability \citep{Yang_Guo_Wang_Xu_2021}. Domain scientists seek to characterize patterns and associations among physical quantities for prediction, uncertainty quantification (UQ), and mechanistic understanding, but they often have only \emph{partial knowledge} of these physical quantities: some factors are measured and their link to the data distribution is known (``known knowns''), others are measured but their influence is uncertain (``known unknowns''), and many remain unanticipated (``unknown unknowns'') \cite{Hatfield2022}. 
%Sensitivity analyses and response-surface studies over physical inputs are central to scientific inference, UQ, and forecasting \cite{razavi2016new, raghavan2018sensitivity}. 
In this setting, DGMs frequently learn \emph{entangled} latent spaces, where a single coordinate influences multiple unrelated aspects of the data, thereby hindering interpretability and downstream applications (e.g., sensitivity analysis, inverse design, hypothesis testing). \emph{Disentangled} representations instead aim for latent coordinates that correspond to independent factors of variation, so that adjusting one coordinate affects only its associated factor \citep{disentangled-review}.  Astronomical imaging provides a compelling testbed; we focus on dark matter halos observed via the thermal Sunyaev--Zel'dovich effect (tSZ, \cite{Zeldovich:1969ff}), identified in maps of the cosmic microwave background (CMB), where images exhibit rich structure tied to physically meaningful, computable auxiliary information, e.g., halo mass and concentration. These auxiliary variables can be computed for each halo map and paired with the image. However, existing unsupervised approaches for disentanglement (e.g., $\beta$-VAE \citep{betavae}, FactorVAE \citep{factor_VAE}, DIP-VAE \citep{DIP_VAE}) only encourage factorization in the latent space, but do not exploit partially known physical covariates like these mass and concentration. When such covariates are available and scientifically important, a pragmatic alternative is to \emph{guide} disentanglement with auxiliary variables, softly steering selected latent coordinates to align with target factors while allowing the remaining latents to capture residual variability.

Recent work \citep{auxvae} proposed \emph{AuxVAE} which has instantiated this idea within a VAE by partitioning the latent space into two segments - an auxiliary-informed block and a residual block, and adding lightweight penalties that (i) align each auxiliary-informed dimension with its corresponding physical variable and (ii) discourage cross-correlation with other latents, while leaving the residual block free to model remaining variation. This improves interpretability without requiring full supervision over all factors. However, across rich, high-detail datasets, including the present application to tSZ halo maps, standard VAEs often over-smooth fine structure, under-represent small-scale variability, and lag behind more expressive generative models in sample fidelity and generalization \citep{vae_failure,vae_rate}.

To overcome these limitations, we turn to \emph{conditional flow matching} (CFM), a class of powerful flow-based DGMs that learn sample-generation by regressing probability flow (or transport) vector fields between simple reference distributions and data distributions \citep{lipman2022flow, tong2023improving}. CFM inherits the benefits of sharpness and diversity from continuous normalizing flows, while enjoying stable and scalable training via supervised regression of vector fields. Recent approaches in flow matching \cite{guo2025variational, samaddar2025efficientflowmatchingusing} have adapted the deep latent variable models (such as VAEs) to the flow matching for structured generation, efficient training, and accurate inference. 
% \AS{Add a sentence about LCFM} 
In this paper, we propose 
% a hybrid architecture, 
\emph{Disentangled Latent-Conditional Flow Matching }(\texttt{DL-CFM}), that marries the interpretability of auxiliary-guided VAEs with the fidelity of CFM. Concretely, we first infer a low-dimensional code $z$ with a VAE encoder and impose the same auxiliary alignment regularizers used in \cite{auxvae} so that selected coordinates of $z$ correspond to known physical factors (e.g. halo mass and concentration). We then train a CFM-based generator conditioned on $z$ to produce high-resolution tSZ maps. In effect, the VAE encoder provides a structured, interpretable bottleneck, and the flow-matching decoder renders those factors into realistic, high-detail images. 
%The disentanglement-enforcing losses are applied in the latent space of the L-CFM model, preserving the intended alignment with auxiliary variables while benefiting from the expressive power of CFM.
In summary, this paper makes the following contributions:
\begin{itemize}
    \item \textbf{Disentanglement in flow matching.} We introduce \texttt{DL-CFM}, bringing auxiliary-variable guidance into conditional flow matching via a lightweight VAE encoder with simple alignment/decoupling losses. To our knowledge, this is the first approach to enable disentangled control within CFM without degrading fidelity.
    \item \textbf{Application in tSZ map generation and control.} On simulated tSZ maps of halos, \texttt{DL-CFM} learns accurate data distribution and generates realistic maps of diverse samples with interpretable control along mass and concentration. Guided traversals allow targeted synthesis at specified settings and separate known factors from residual morphology.
    \item \textbf{Scientific validation and diagnostics.} The learned latents recover the expected mass-concentration trend and surface outliers (e.g., disturbed systems or artifacts), enabling sensitivity analyses and
    anomaly discovery with a compact, interpretable representation, as well as fast generation of realistic mock datasets from minimal inputs.
\end{itemize}
Next, we describe our approach in Section~\ref{sec:method} and present experimental results in Section~\ref{sec:results}.

\section{Methodology}\label{sec:method}
We use auxiliary physical variables to shape the latent space of a deep generative model so that selected coordinates align with known factors (e.g., halo mass and concentration) while preserving generative flexibility. Concretely, we propose \texttt{DL-CFM}, which adapts the AuxVAE loss function to the state-of-the-art latent conditional flow matching model for high-quality sample generation and physics-aware latent space disentanglement.
% (i) build an auxiliary-variable–guided VAE (AuxVAE) that provides an interpretable, physics-aware latent bottleneck and (ii) feed this bottleneck into a high-fidelity generator trained via latent–conditional flow matching (L-CFM). 
This section specifies the setting and the \emph{main loss terms} needed to reproduce our method; extended definitions and derivations are deferred to the Appendix.

\subsection{Notations}
Let $x\!\in\!\mathbb{R}^{p}$ be an observation (a tSZ halo image) and $u\!\in\!\mathbb{R}^{d}$ the auxiliary variables (here, halo mass and concentration). The VAE introduces a latent $z\!\in\!\mathbb{R}^{d_Z}$ and partitions it as
\[
z=\big(z_{\mathrm{aux}},\,z_{\mathrm{rec}}\big),\qquad
z_{\mathrm{aux}}\in\mathbb{R}^{d}\ \text{(auxiliary-guided)},\quad
z_{\mathrm{rec}}\in\mathbb{R}^{d_Z-d}\ \text{(reconstruction-focused)}.
\]
To align $z_{\mathrm{aux}}$ with $u$ while leaving $z_{\mathrm{rec}}$ free to capture remaining “unknown unknowns,” we use an \emph{auxiliary-informed prior}
\begin{equation}
\label{eq:prior_main}
p(z\mid u)=\mathcal{N}\!\left(\mu_0(u),\Sigma_0\right),\quad
\mu_0(u)=\big(u_1,\dots,u_d,\,0,\dots,0\big),\quad
\Sigma_0=\mathrm{diag}\!\left(\tau^2\mathbf{I}_{d},\,\mathbf{I}_{d_Z-d}\right),
\end{equation}
where $\tau^2\!\ll\!1$ is a small variance, generally taken as inverse of the batch-size, that softly tethers the guided coordinates to $u$ (we normalize $u$ to $[0,1]$).

\subsection{Auxiliary-variable guided disentangled Latent CFM}
Recent approaches in flow matching \cite{guo2025variational, samaddar2025efficientflowmatchingusing} have combined deep latent variable models with flow-based generative models to ensure efficient training and inference. We propose \texttt{DL-CFM} 
that extends these state-of-the-art flow matching approaches to enforce disentanglement in the latent space using the auxiliary variables in our data. 

% With encoder $q_\phi(z\mid x)$ and decoder $p_\theta(x\mid z)$, the AuxVAE maximizes
We aim to learn a time-dependent vector field $v_\theta (x_t,z,t)$ that evolves samples from a simple source distribution, $x_0$, to the high-dimensional halo dataset conditioned on the disentangled latent variable $z$. To enforce the disentanglement, we propose the loss function, 
\begin{align}
\label{eq:auxvae_mainloss}
\mathcal{L}_{\texttt{DL-CFM}}
= & \underbrace{\E_{p(t),q_\phi(z|x),p_t(x_t|x_0,x)}   ||v_\theta (x_t,z,t)- u_t(x_t|x_0,x_1)||_2^2}_{\text{conditional flow matching loss}}
+\beta \underbrace{\mathrm{KL}\!\left(q_\phi(z\mid x)\,\|\,p(z\mid u)\right)}_{\text{conditional prior match}}  \nonumber\\
& +\lambda_{1}\!\sum_{j=1}^{d}\!\Big(\underbrace{\mathsf{Align}\!\left(u_j,\mu_{\phi,\mathrm{aux},j}\right)}_{\text{explicitness}}
+\underbrace{\mathsf{Decorr}\!\big(u_j,\mu_{\phi,\mathrm{aux},-j}\big)}_{\text{intra-independence}}\Big)
+\lambda_{2}\,\underbrace{\mathsf{Decorr}\!\big(u,\mu_{\phi,\mathrm{rec}}\big)}_{\text{inter-independence}}.
\end{align}
% \begin{align}
% \label{eq:auxvae_mainloss}
% \mathcal{L}_{\texttt{DL-CFM}}
% = & \underbrace{\mathbb{E}_{q_\phi(z\mid x)}\!\left[\log p_\theta(x\mid z)\right]}_{\text{reconstruction}}
% -\underbrace{\mathrm{KL}\!\left(q_\phi(z\mid x)\,\|\,p(z\mid u)\right)}_{\text{conditional prior match}}  \nonumber\\
% & -\lambda_{1}\!\sum_{j=1}^{d}\!\Big(\underbrace{\mathsf{Align}\!\left(u_j,\mu_{\phi,\mathrm{aux},j}\right)}_{\text{explicitness}}
% +\underbrace{\mathsf{Decorr}\!\big(u_j,\mu_{\phi,\mathrm{aux},-j}\big)}_{\text{intra-independence}}\Big)
% -\lambda_{2}\,\underbrace{\mathsf{Decorr}\!\big(u,\mu_{\phi,\mathrm{rec}}\big)}_{\text{inter-independence}}.
% \end{align}
Here $\mu_{\phi}(\cdot)$ denotes the encoder mean; $\mu_{\phi,\mathrm{aux}}$ and $\mu_{\phi,\mathrm{rec}}$ are its restrictions to the auxiliary-guided and reconstruction-focused coordinates, and $\mu_{\phi,\mathrm{aux},-j}$ excludes the $j^{\text{th}}$ guided coordinate. The three regularizers enforce: (i) \emph{explicitness} (guided coordinate $j$ tracks $u_j$ in a one-to-one manner), (ii) \emph{intra-independence} (no cross-correlation among guided latents), and (iii) \emph{inter-independence} (reconstruction-focused latents are decorrelated from $u$). We instantiate $\mathsf{Align}$ and $\mathsf{Decorr}$ as lightweight correlation-based penalties computed from minibatch statistics of $\mu_{\phi}$. We leave the details of training and sampling for \texttt{DL-CFM}
% precise definitions and alternatives are provided 
in App.~\ref{app:auxvae_details}. Equation~\eqref{eq:auxvae_mainloss} introduces no extra networks and only a few scalar hyperparameters $(\beta, \lambda_1,\lambda_2,\tau^2)$.

% \paragraph{Practical recipe.}
% (i) Select the $d$ guided factors in $u$ and reserve $d$ latent coordinates for $z_{\mathrm{aux}}$; (ii) train with \eqref{eq:prior_main} and \eqref{eq:auxvae_mainloss}; (iii) use the learned $(z_{\mathrm{aux}},z_{\mathrm{rec}})$ as a structured bottleneck to drive a high-fidelity generator (next subsection).

% \subsection{Disentangled L-CFM} \textbf{\textcolor{red}{Anirban}}

\section{Experimental results} \label{sec:results}
We evaluate (i) \emph{generation quality} and (ii) \emph{disentanglement effects} on synthetic thermal Sunyaev--Zel'dovich (tSZ) halo images. Our model of interest is \texttt{DL-CFM}, which couples an auxiliary-guided VAE bottleneck to a conditional flow matching model. We compare \texttt{DL-CFM} to a state-of-the-art baseline ICFM \cite{tong2023improving} model in terms of generation quality.

\subsection{Data and experimental setup}
\textbf{Simulations and halos.}
We use cosmological hydrodynamic simulations from \cite{2025OJAp....8E..82D} run with the Conservative Reproducing Kernel, Hybrid/Hardware Accelerated Cosmology Code (CRK-HACC; \cite{2023ApJS..264...34F}); details are in App.~\ref{app:data}. Halos are identified with a friends-of-friends finder, and halo centers are set to the most bound dark matter particle. For each halo, we compute two physical properties that are expected to capture key intracluster medium (ICM) morphology \cite{2024OJAp....7E.116K}: mass $M_{200c}$ and concentration $c_{200c}$. We select halos with $M_{200c}>10^{13.5}\,h^{-1}M_\odot$. Details on the halo catalog and associated properties are detailed in Appendix \ref{app:data_simulation}. 
We represent each halo with a Compton-$y$ map—the line-of-sight integral of the electron pressure—of the thermal Sunyaev–Zel’dovich (tSZ) signal. Each image is paired with its $(M_{200c}, c_{200c})$ values. Our experiments test whether \texttt{DL-CFM} matches the sample fidelity of CFM \emph{while enabling controlled generation} along the mass and concentration axes.

\subsection{Generation quality}
% \textbf{\textcolor{red}{Anirban}}
\begin{table}
    \centering
    \resizebox{0.9\textwidth}{!}{
    \begin{tabular}{|c|c|c|c|c|c|}
    \hline
        % &\multicolumn{2}{c|}{\textbf{I-CFM}} & \multicolumn{3}{c|}{\texttt{DL-CFM}}  \\
        % \hline
      \textbf{Methods} & \textbf{\# Params.}  & \textbf{Sinkhorn} ($\downarrow$) & \textbf{Energy} ($\downarrow$) & \textbf{Gaussian} ($\downarrow$) & \textbf{Laplacian} ($\downarrow$) \\
      \hline
     ICFM  & 34.42M & \textbf{4564.03} $\pm$ 37.883 & 84.073 $\pm$ 0.797 & \textbf{0.00813} $\pm$ 0.00020 & 0.00693 $\pm$ 0.00011\\
     \hline
     \texttt{DL-CFM}  & 38.06M & 4819.211 $\pm$ 32.718 & \textbf{83.148} $\pm$ 0.767 & \textbf{0.00813} $\pm$ 0.00014 & \textbf{0.00678} $\pm$ 0.00012\\
     \hline
    \end{tabular}
    }
    \caption{Table shows the generation quality for ICFM and \texttt{DL-CFM} in terms of different distance metrics (mean $\pm$ Sd). In terms of most metrics, the two approaches show similar generation quality.
    % , with \texttt{DL-CFM} performing marginally better in terms of Energy distance.
    }
    \label{tab:generation_quality}
\end{table}
Table~\ref{tab:generation_quality} shows different distance metrics (mean $\pm$ sd) calculated between the training samples and the generated samples from ICFM and \texttt{DL-CFM} model trained on the tSZ halo data set. We observe that, in terms of most distance metrics, both approaches show similar generation quality. \texttt{DL-CFM} performs marginally better in terms of the Energy metric. Both methods show a large Sinkhorn distance, with ICFM showing a lower distance than \texttt{DL-CFM}.  
% \begin{table*}
% \centering
% \resizebox{\textwidth}{!}{   
% \begin{tabular}{|c|c|c|c|c|c|c|c|c|}
%    \hline
%    \multirow{3}{1.5cm}{\textbf{Methods}}  & \multicolumn{4}{c|}{\textbf{CIFAR10}}  &  \multicolumn{4}{c|}{\textbf{MNIST}}\\
%    \cline{2-5} \cline{6-9} & \multirow{2}{1.5cm}{\textbf{\# Params.}} & \multicolumn{3}{c|}{\textbf{FID ($\downarrow$)}}  & \multirow{2}{1.5cm}{\textbf{\# Params.}} & \multicolumn{3}{c|}{\textbf{FID ($\downarrow$)}}\\
%    \cline{3-5} \cline{7-9}
%    & & \textbf{100} & \textbf{1000} & \textbf{Adaptive} & & \textbf{100} & \textbf{1000} & \textbf{Adaptive}\\
% 	\hline
%     VRFM-1 \cite{guo2025variational} &37.2 M & 4.349& 3.582 & 3.561& - & - & - & - \\
%     VRFM-2 \cite{guo2025variational} &37.2 M& 4.484 & 3.614 & \textbf{3.478} & - & - & - & - \\
%     \hline
% 	OT-FM & 35.8 M & 4.661  & 3.862  & 3.727 & 1.56 M & 15.101  & 15.880  & 16.012   \\
%     I-CFM & 35.8 M & 4.308  & \textbf{3.573}  & 3.561 & 1.56 M & 14.272  & 14.928  & 15.050   \\
%     \hline
% 	\texttt{Latent-CFM} & 36.1 M & \textbf{4.246}  & 3.575  & 3.514  & 1.58 M & \textbf{13.848}  & \textbf{14.543}  & \textbf{14.694} \\
%     % \texttt{Latent-CFM} (large) & 34.5 M & 4.246 \semitransp{(0.0)} & 3.575 \semitransp{(0.0)} & 3.514 \semitransp{(0.0)} \\
%    \hline
%    \end{tabular}
%    }
%    \caption{}
%     \label{tab:generation_quality}
% \end{table*}

\subsection{Disentanglement effects}
We test whether the guided latents isolate the intended auxiliary information and support controlled generation. \textbf{(a) Latent–auxiliary alignment:}
We scatter $\{\log M_{200c},\,c_{200c}\}$ against the first five latent coordinates: the first two are auxiliary-guided ($z_{\mathrm{aux}}$) and the next three are representative reconstruction-focused ($z_{\mathrm{rec}}$). \texttt{DL-CFM} shows near one-to-one, monotonic relationships on the guided axes and weak correlations elsewhere, consistent with the training objective; the generated mass–concentration trend matches the simulation catalog (Fig.~\ref{fig:z_vs_u}). \textbf{(b) Controlled traversals.}
We traverse each guided dimension in $z_{\mathrm{aux}}$ while holding $z_{\mathrm{rec}}$ fixed (Fig.~\ref{fig:traversal}). Samples vary systematically and interpretably along mass and concentration, enabling targeted generation of tSZ halos at desired $(M_{200c},c_{200c})$ without sacrificing fidelity.

% preamble: \usepackage{capt-of}  % or \usepackage{caption}
\begin{center}
\begin{minipage}{0.48\linewidth}
  \centering
  \includegraphics[width=\linewidth]{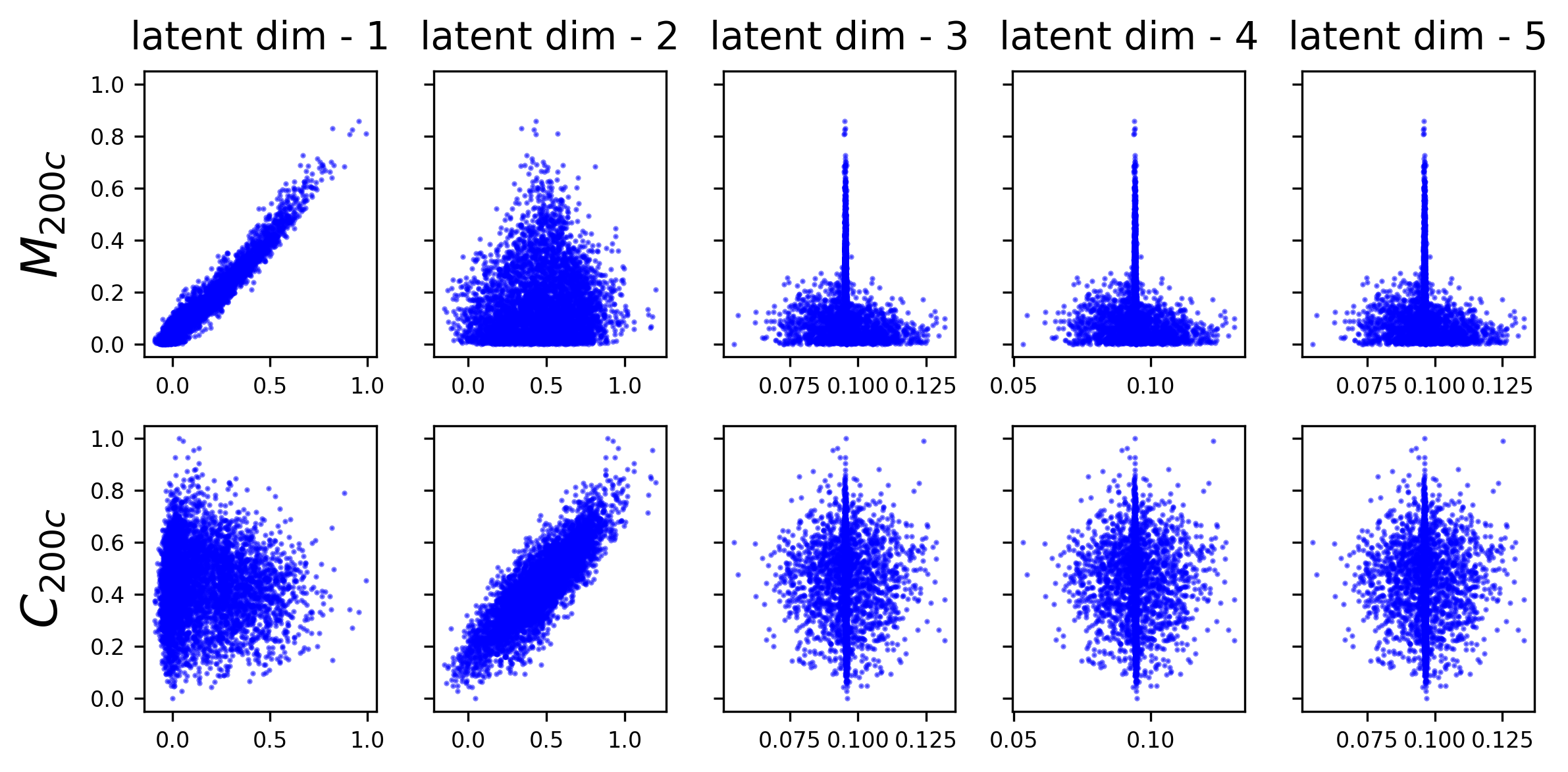}
  \captionof{figure}{Alignment of guided latents with mass and concentration.}
  \label{fig:z_vs_u}
\end{minipage}\hfill
\begin{minipage}{0.48\linewidth}
  \centering
  \includegraphics[width=\linewidth]{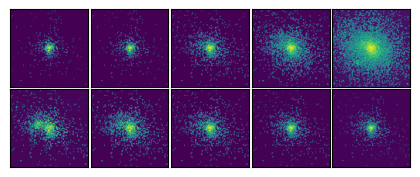}
  \captionof{figure}{Traversals along guided latents ($z_{\mathrm{aux}}$) with $z_{\mathrm{rec}}$ fixed. Rows: mass, concentration.}
  \label{fig:traversal}
\end{minipage}
\end{center}

% One figure number, two stacked images, single caption at the bottom
\begin{figure}[!h]
  \centering
  \includegraphics[width=0.95\linewidth]{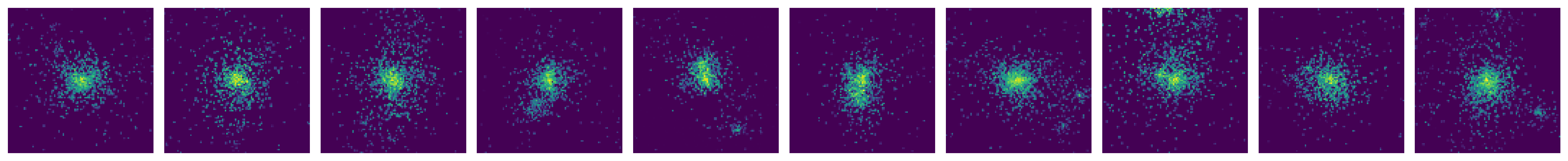}

  \medskip

  \includegraphics[width=0.95\linewidth]{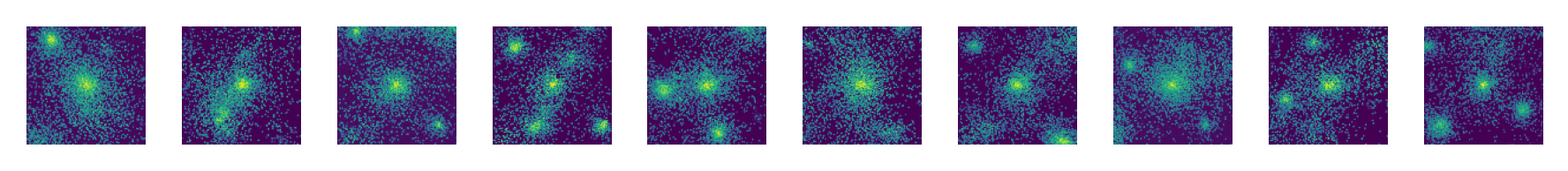}

  \caption{Generating samples from the center (top) and tail (bottom) of the reconstruction-focused latents $z_{\mathrm{rec}}$, with the first two auxiliary-guided coordinates fixed at $(z_1,z_2)=(0.001,0.001)$.}
  \label{fig:center_tail_recon}
\end{figure}

\paragraph{(c) Conditional distributional shifts.}
We illustrate controlled generation by fixing the auxiliary-guided latents to a low-mass, low-concentration setting, $(z_1,z_2)=(0.001,0.001)$ (with $z_1$ and $z_2$ aligned to mass and concentration), and sampling only the reconstruction-focused latents $z_{\mathrm{rec}}$. Figure~\ref{fig:center_tail_recon} shows generated tSZ halos from two regimes: the \emph{center} of the $z_{\mathrm{rec}}$ distribution (top) and its \emph{tails} (bottom). Center samples appear relaxed and single-peaked, whereas tail samples exhibit complex, multi-peaked morphologies indicative of disturbed systems or active merger status. This conditional shift demonstrates that $z_{\mathrm{rec}}$ captures residual structure beyond $(M_{200c},c_{200c})$ and enables targeted sampling for sensitivity analyses and discovery—e.g., identifying cases where center-based concentration estimates under-represent complex, multi-core systems. Additional settings (high-mass/low-concentration and low-mass/high-concentration) are shown in Appendix \ref{sec:additional_results}.

\section{Discussion}
\texttt{DL-CFM} couples auxiliary-guided latents with conditional flow matching to synthesize tSZ halo maps with high fidelity and interpretable control. Simple alignment and decorrelation penalties expose known factors while allowing residual latents to capture remaining unknown variability; these latents generate diverse structures under fixed auxiliary settings. Overall, auxiliary-guided flows offer a compact route to uniting interpretability with state-of-the-art generative quality. Applying the method to real data will require careful selection and calibration of auxiliary variables and explicit treatment of instrumental or systematic effects.
\begin{ack}
This material is based upon work supported by the U.S. Department of Energy, Office of Science, Office of Advanced Scientific Computing Research (ASCR) and the Scientific Discovery through Advanced Computing (SciDAC) Program Nuclear Physics partnership titled ``Femtoscale Imaging of Nuclei using Exascale Platforms'' and the SciDAC-RAPIDS institute, FASTMath Institute programs under Contract No. DE-AC02-06CH11357. Work at Argonne National Laboratory is supported by UChicago Argonne LLC, Operator of Argonne National Laboratory. Argonne, a U.S. Department of Energy Office of Science Laboratory, is operated under Contract No. DE-AC02-06CH11357. The training is carried out on Swing, a GPU system at the Laboratory Computing Resource Center (LCRC) of Argonne National Laboratory.
\end{ack}

%\section*{References}
\bibliographystyle{unsrtnat}  % or abbrvnat, unsrtnat
\bibliography{bib}

%%%%%%%%%%%%%%%%%%%%%%%%%%%%%%%%%%%%%%%%%%%%%%%%%%%%%%%%%%%%

\appendix

% \section{Technical Appendices and Supplementary Material}
% Technical appendices with additional results, figures, graphs and proofs may be submitted with the paper submission before the full submission deadline (see above), or as a separate PDF in the ZIP file below before the supplementary material deadline. There is no page limit for the technical appendices.
\renewcommand{\thefigure}{A.\arabic{figure}}
\renewcommand{\thetable}{A.\arabic{table}}
\renewcommand{\thealgorithm}{A.\arabic{algorithm}}
\setcounter{figure}{0}
\setcounter{table}{0}
\setcounter{algorithm}{0}

\section{Data}\label{app:data}

To train the proposed \texttt{DL-CFM} model, we use synthetic images produced from a cosmological hydrodynamic simulation.
The set of simulations we use is described in detail in \cite{2025OJAp....8E..82D}; in this section, we briefly summarize their main properties and the generation of the images.

\subsection{Simulations}\label{app:data_simulation}
The simulation set was generated using the Hybrid/hardware Accelerated Cosmology Code (HACC, \cite{2016NewA...42...49H}); more specifically, we focus on simulations produced using the CRK-HACC hydrodynamic solver described in \cite{2023ApJS..264...34F}.
Initial conditions are generated shortly after the Big Bang, at redshift $z=200$, using the best-fit cosmology derived from the \textit{Planck} analysis of the cosmic microwave background (\cite{2020A&A...641A...6P}).
The simulation spans a volume of $(576 \; h^{-1} {\rm Mpc})^3$ (comoving) and simulates the evolution of $2 \times (2304)^3$ particles (half dark matter, half gas) all the way to present day, at $z=0$.
We specifically focus on the ``non-radiative'' (or ``adiabatic'') version of the hydrodynamic simulation (\cite{2025OJAp....8E..82D}), in which all baryonic matter is modeled as gas interacting through hydrodynamic equations, and no sub-resolution physics are included.

\subsection{Halo catalog}
In this work, we focus on the very last stage of the simulation (present day, $z=0$).
A friends-of-friends (FoF) algorithm is run on the dark matter particles of the simulation to identify halos.
For each halo found by the FoF finder, the center of the halo is defined as the position of the most bound dark matter particle, and the entire matter distribution (both gas and dark matter) around this position are used to compute halo properties.

In this work, we focus on a small subset of halo properties, expected to contain most of the information about a halo and its gas distribution \cite{2024OJAp....7E.116K}:
\begin{enumerate}
    \item Halo mass $M_{200c}$, computed as the mass enclosed within a characteristic radius, $R_{200c}$, within which the average halo matter density is 200 times greater than the critical density of the Universe;
    \item Halo concentration $c_{200c}$, quantifying how much of the halo matter is contained in the center of the halo as opposed to its outskirts;
    \item Distance between the halo potential peak and center of mass, $\Delta x / R_{200c}$, known to be an indicator of the disturbed state of halos (specifically of their merging status, see \textit{e.g.} \cite{2018ApJ...859...55C};
    \item Halo ellipticity, $e$, and prolaticity, $p$, quantifying the overall triaxial shape of the halo.
\end{enumerate}

\subsection{Image generation}
The thermal Sunyaez-Zel'dovich is a spectral distortion of the Cosmic Microwave Background due to its interaction with free electrons, in particular in the hot plasma forming the intracluster medium (ICM) of massive halos.
The amplitude of this distortion is given by the Compton-$y$ parameter, proportional to the line-of-sight (LoS) integral of the electron pressure $P_{\rm e}$ in the ICM:
\begin{equation}
    y = \frac{\sigma_{\rm T}}{m_{\rm e} c^2} \int_{\rm LoS} P_{\rm e} \, {\rm d}l,
    \label{eq:ysz}
\end{equation}
where $\sigma_{\rm T}$ is the Thompson scattering cross-section, and $m_{\rm e} c^2$ is the electron rest-frame energy.

The images used in this work are maps of the Compton-$y$ parameter in massive ($M_{200c} > 10^{13.5} \, h^{-1} M_\odot$) dark matter halos.
For each halo, the gas density and temperature is projected on a 3D grid using the cloud-in-cell algorithm.
The grid is $(64)^3$ cells, with a box size of $(4 \times R_{200c})^3$.
The gas pressure is computed as the product of density and pressure, and converted to electron pressure using the ratio of the mean molecular weight of the fully ionized gas and of electrons.
The resulting 3D pressure distribution is then integrated along three orthogonal lines of sight using eq.~\ref{eq:ysz}, resulting in three $(64\times64)$ Compton-$y$ images per halo.

\section{\texttt{DL-CFM} details and derivations}
\label{app:auxvae_details}
\subsection{Flow matching details}
Flow matching attempts to transport samples $x_0$ from a simple source distribution with density $p_0$ to a complex data distribution with density $p_1$. This is done through a time-dependent vector field $u_t:[0,1]\times \mathbb{R}^p \rightarrow \mathbb{R}^p$ which defines an ordinary differential equation, 
\begin{equation}\label{eq:vec_field}
    \frac{d \phi_t(x)}{dt} = u_t(\phi_t(x)); \phi_0(x) = x_0
\end{equation}
where $\phi_t(x) = x_t$ is the solution of the ODE or \textit{flow} with the initial condition in Eq~\ref{eq:vec_field}. The primary objective of flow matching models is to train a neural network $v_\theta(.,t)$ to learn the ground-truth vector field $u_t$. However, the true vector field is unknown for most real-world datasets. Therefore, the common approach is to fix the conditional vector field $u_t(.|x_0,x)$ conditioned on the sample pairs $(x_0,x) \sim q(x_0,x)$, where $q$ is a joint distribution with marginals $p_0$ and $p_1$. Following the ICFM \cite{tong2023improving}, we fix $u_t(.|x_0,x) = x-x_0$ and $q = p_0 \times p_1$ and $p_0 = N(0,I)$ as the source distribution. This leads to the conditional flow matching objective,
\begin{equation}\label{eq:cond_flow}
    \mathcal{L}_{\text{CFM}} = \E_{p(t),q(x_0,x),p_t(x_t|x_0,x)} \left[ ||v_\theta(x_t,t) - u_t(x_t|x_0,x_1)||_2^2 \right]
\end{equation}
where $p_t(.|x_0,x)$ is the probability path which we fix to be $N(.; tx + (1-t)x_0,\sigma^2I_p)$. Following Theorem 2.1 in \cite{tong2023improving}, one can show that the fixed vector field $u_t(.|x_0,x) = x-x_0$ induces this probability path. With $p(t) = unif(0,1)$, Eq.~\ref{eq:cond_flow} is a tractable objective which can be minimized with respect to the neural network parameters $\theta$.

Recent approaches in \cite{samaddar2025efficientflowmatchingusing} have extended the CFM model to incorporate data-driven latent structures. The authors model the data as a latent mixture model governed by the latent variable $z$, $p_1 =  \int q (z) q(.|z) dz$. The latent variables are learned from the data using a pretrained latent variable model. In this work, we learn the latent variables from the data using a lightweight encoder model. Instead of pretraining, we train the encoder along with the learned vector field parameters $\theta$, maximizing the \texttt{DL-CFM} loss in Eq.~\ref{eq:auxvae_mainloss}.

\subsection{ELBO with conditional prior and closed-form KL}
With a Gaussian encoder $q_\phi(z\mid x)=\mathcal{N}\!\big(\mu_\phi(x),\Sigma_\phi(x)\big)$ and the prior in \eqref{eq:prior_main}, the per-sample KL term admits the closed form
\begin{align}
\mathrm{KL}\!\left(q_\phi(z\mid x)\,\|\,p(z\mid u)\right)
&=\tfrac{1}{2}\!\left[
\log\frac{|\Sigma_0|}{|\Sigma_\phi|}-d_Z
+\big(\mu_\phi-\mu_0\big)^\top\Sigma_0^{-1}\big(\mu_\phi-\mu_0\big)
+\mathrm{tr}\!\big(\Sigma_0^{-1}\Sigma_\phi\big)
\right].
\end{align}
The reconstruction term is Monte Carlo–estimated via samples from $q_\phi(z\mid x)$.

\subsection{Why regulate the expected variational posterior}
Let $q_\phi(z)=\!\int q_\phi(z\mid x)p(x)\,dx$ and $p_\theta(z)=\!\int p_\theta(z\mid x)p(x)\,dx$. By convexity of $\mathrm{KL}$ \citep{DIP_VAE},
\begin{align}
\label{eq:kl_variational_posterior}
\mathrm{KL}\!\big(q_\phi(z)\,\|\,p_\theta(z)\big)
=\mathrm{KL}\!\Big(\mathbb{E}_{p(x)}q_\phi(z\mid x)\,\Big\|\,\mathbb{E}_{p(x)}p_\theta(z\mid x)\Big)
\ \le\
\mathbb{E}_{p(x)}\mathrm{KL}\!\big(q_\phi(z\mid x)\,\|\,p_\theta(z\mid x)\big).
\end{align}
Maximizing the standard ELBO decreases the RHS of \eqref{eq:kl_variational_posterior} but may still leave residual dependencies in $q_\phi(z)$. Our correlation regularizers directly target these dependencies at the \emph{population} level using minibatch estimates.

\subsection{Correlation-based regularizers}
Let $v\in\mathbb{R}^{m_v}$ and $w\in\mathbb{R}^{m_w}$. Define
\[
\Sigma_{vw}=\mathbb{E}\!\left[(v-\mathbb{E}v)(w-\mathbb{E}w)^\top\right],\qquad
\mathrm{Corr}(v,w)=\mathrm{diag}(\Sigma_{vv})^{-\frac{1}{2}}\ \Sigma_{vw}\ \mathrm{diag}(\Sigma_{ww})^{-\frac{1}{2}}.
\]
To capture nonlinear relations we use polynomial lifts up to degree $K$ (applied elementwise), and aggregate:
\begin{align}
\label{eq:R0}
R_0^{K}(v,w)&=\frac{1}{K\,m_v m_w}\sum_{k\neq k'}\sum_{i=1}^{m_v}\sum_{j=1}^{m_w}
\left|\mathrm{Corr}\!\left(v^{k},w^{k'}\right)_{ij}\right|,
\\
\label{eq:R1}
R_1^{K}(v,w)&=\frac{1}{K\,m_v m_w}\sum_{k\neq k'}\sum_{i=1}^{m_v}
\Big(1-\left|\mathrm{Corr}\!\left(v^{k},w^{k'}\right)_{ii}\right|\Big).
\end{align}
Intuition: $R_0$ penalizes generic cross-dependence (off-diagonals), while $R_1$ rewards one-to-one alignment (diagonals close to $\pm1$).

\paragraph{Surrogate with encoder means.}
Let $\mu_\phi(x)=\mathbb{E}[z\mid x]$. For polynomials $k,k'$, by the law of total variance,
\begin{align}
\mathrm{Cov}\!\big(u^{k},z_{\mathrm{rec}}^{k'}\big)
&=\mathbb{E}_{(x,u)}\Big[\underbrace{\mathrm{Cov}\!\big(u^{k},z_{\mathrm{rec}}^{k'}\mid x,u\big)}_{=\,0}\Big]
+\mathrm{Cov}_{(x,u)}\!\Big(u^{k},\,\mathbb{E}[z_{\mathrm{rec}}^{k'}\mid x]\Big)
=\mathrm{Cov}_{(x,u)}\!\big(u^{k},\,\mu_{\phi,\mathrm{rec}}^{k'}\big),
\end{align}
and analogously for the guided block. Hence, batch correlations of $\mu_\phi$ suffice to estimate dependencies.

\subsection{Instantiating \texorpdfstring{$\mathsf{Align}$}{Align} and \texorpdfstring{$\mathsf{Decorr}$}{Decorr}}
We use
\[
\mathsf{Align}(u_j,\mu_{\phi,\mathrm{aux},j}) \ =\ 1- R_1^{K}(u_j,\mu_{\phi,\mathrm{aux},j}),\qquad
\mathsf{Decorr}(a,b) \ =\ R_0^{K}(a,b),
\]
which produces the main loss in \eqref{eq:auxvae_mainloss}. Both terms are scale-free and computed from standardized batch statistics; $K\!\in\!\{1,2\}$ works well in practice.

\subsection{Training and inference algorithms}
Algorithm~\ref{alg:vae-cond} shows the training steps of \texttt{DL-CFM}. Given $n$ images, the regularizers, and the initialized networks, we draw a single sample $z_i$ from the encoder distribution $q_\phi (.|x_i)$ for each $x_i$. These are concatenated with the noisy samples $x^{t_i}_i$ and the noise level $t_i$ and passed through the vector field network $v_\theta(.,.,.)$. The latent variables, the output of the vector field network, and the true conditional vector field target are used in the disentangled loss in Eq.~\ref{eq:auxvae_mainloss}, which is optimized with respect to the parameters $\theta, \phi$.

\begin{algorithm}
\caption{\texttt{DL-CFM} training}\label{alg:vae-cond}
\begin{algorithmic}[1]
\STATE Given $n$ sample $(x_1, ..., x_n)$ from $p_1(.)$, regularizers $\beta, \lambda_1, \lambda_2$
% \IF{no pretrained VAE available}
% \STATE Train \texttt{VAE} using $(x^1_1, ..., x^1_n)$ optimizing Eq.~\ref{eq:loss_VAE}
% \STATE Save the encoder $q_{\hat{\lambda}}(.|x_1)$ 
% \ENDIF
% \STATE $\hat{\Q} \leftarrow$ \texttt{REMEDI}$(\Q)$.
\STATE Initialize $v_{\theta}(\cdot, \cdot,\cdot)$ and encoder layer parameters $\phi$
\FOR{$k$ steps}
\STATE Sample latent variables $z_i \sim q_{\phi}(.|x_i)$ for all $i=1,...,n$
\STATE Sample $(x^0_1, ..., x^0_n)$ from $\mathcal{N}(0, I)$ and noise levels $(t_1,...,t_n)$ from $Unif(0,1)$ and compute $(u_{t_1}(.|x_0,x),...,u_{t_n}(.|x_0,x))$
\STATE compute $v_{\theta}(x_i^{t_i}, z_i,t_i)$ where $x_i^{t_i}$ is the corrupted $i$-th data at noise level ${t_i}$
\STATE Compute $\nabla \mathcal{L}_{\texttt{DL-CFM}}$ and update $\theta, \phi$
\ENDFOR
%\STATE Draw $m$ samples $(\Tilde{x}_1, ..., \Tilde{x}_m)$ from KNIFE
\STATE \textbf{return} $v_{\hat{\theta}}(\cdot, \cdot,\cdot), q_{\hat{\phi}}(.|x)$
\end{algorithmic}
\end{algorithm}

\begin{figure}
    \centering
    \includegraphics[width=\linewidth]{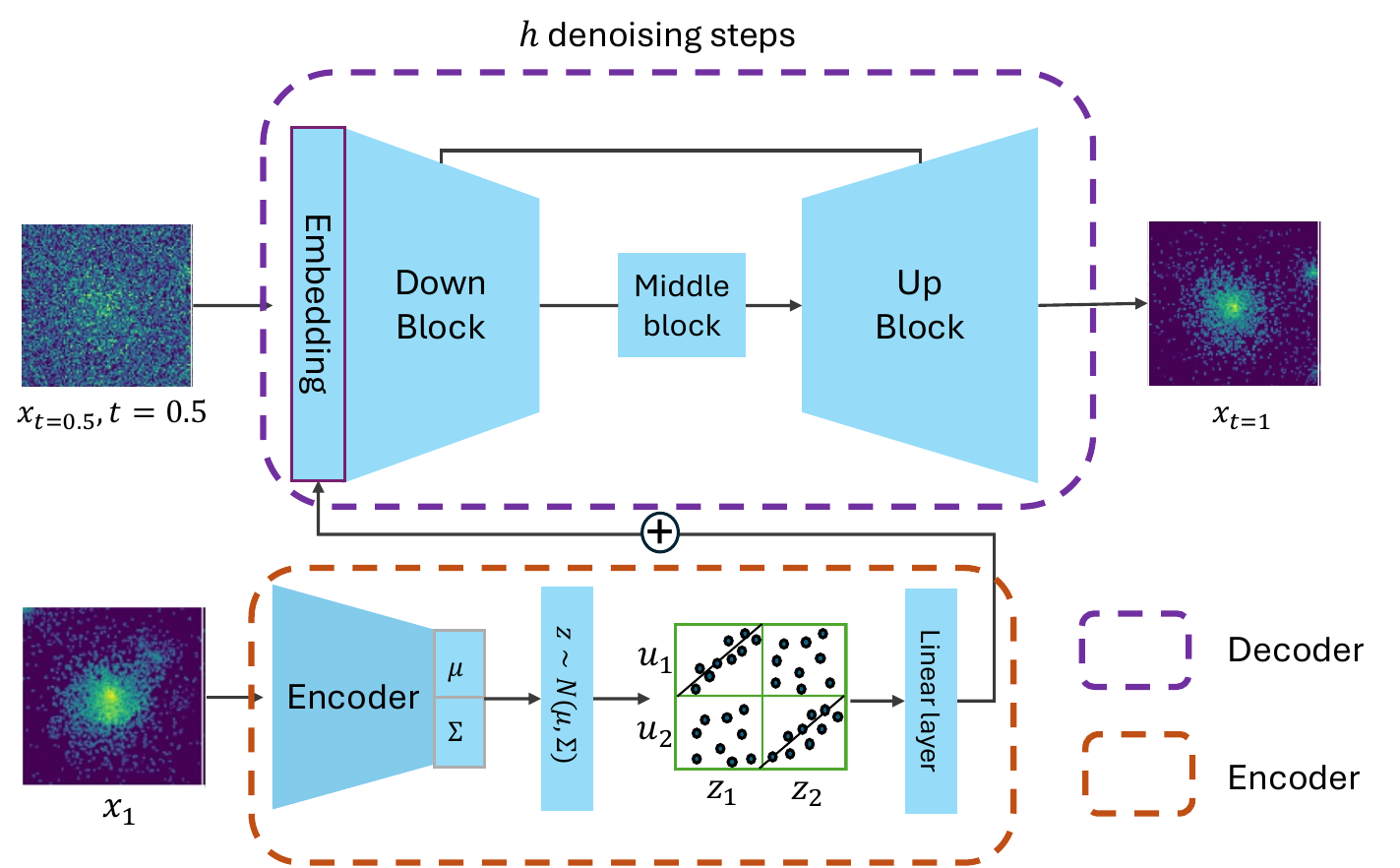}
    \caption{Schematic of the \texttt{DL-CFM} inference. Given a training sample, we fix the sampled latent from the disentangled latent space. The latent variable is used in the vector field network to evolve the source samples to the data distribution. For demonstration, we show two snapshots of the iterative reverse process at $t=0.5$ (left) and $t=1$ (right) using the vector field U-Net.}
    \label{fig:schematic}
\end{figure}

Algorithm~\ref{alg:vae-cond_inference} shows the inference procedure for \texttt{DL-CFM}. Following \cite{samaddar2025efficientflowmatchingusing}, we reuse the training data to draw samples from \texttt{DL-CFM}. Given the budget of $K$ samples, a random batch of $K$ training data is sampled, then passed through the encoder to draw samples from the latent space. For each latent sample $z_i$, we iteratively solve the ODE in Eq.~\ref{eq:vec_field} for $h$ steps.  Fig.~\ref{fig:schematic} shows the schematic of the \texttt{DL-CFM} inference procedure.  Note that the latent sampling is performed once during the inference and fixed for $h$ ODE denoising steps.

\begin{algorithm}
\caption{\texttt{DL-CFM} inference}\label{alg:vae-cond_inference}
\begin{algorithmic}[1]
\STATE Given sample size $K$, trained $v_{\hat{\theta}}(.,.,.)$ and $q_{\hat{\phi}}(.|x)$, number of ODE steps $n_{ode}$
\STATE Select $K$ training samples $(x_1^{train},...,x_K^{train})$
\STATE Sample latent variables $z_i \sim q_{\hat{\phi}}(.|x_i^{train})$ for all $i=1,...,K$
\STATE Sample $(x^0_1, ..., x^0_n)$ from $\mathcal{N}(0, I)$
\STATE $h \leftarrow \frac{1}{n_{ode}}$ 
\FOR{$t=0,h,...,1-h$ and $i=1,...,K$}
\STATE $x_i^{t+h} =$ ODEstep($v_{\hat{\theta}}(x_i^t,z_i,t),x_i^t$)
\ENDFOR
\STATE \textbf{return} Samples $(x_1,...,x_K)$
\end{algorithmic}
\end{algorithm}

\subsection{Implementation details}
% We train with Adam, estimate $\mathbb{E}_{q_\phi(z\mid x)}[\log p_\theta(x\mid z)]$ via one reparameterized sample per input, and compute the correlation penalties from encoder means on each minibatch. Hyperparameters $(\lambda_1,\lambda_2)$ are small (e.g., $10^{-3}$–$10^{-2}$) to preserve reconstruction fidelity while encouraging alignment and (de)correlation. The guided dimensionality $d$ equals the number of auxiliary factors.
The models trained have the same U-Net architecture from \cite{tong2023improving}. For I-CFM, the model takes the input $(x_t,t)$, the variables are projected onto an embedding space and concatenated along the channel dimension, then passed through the U-Net layers to output the learned vector field.

In \texttt{DL-CFM}, we use a deep convolutional neural network as the encoder network. The network consists of four convolutional downsampling blocks, where each convolutional layer is followed by a batch normalization and a leaky ReLU activation. The output is then passed through two dilated convolution blocks with batch normalization and leaky ReLU activation. The output from the convolutional encoder is flattened and passed through a linear layer to predict the mean and the log-variance of the latent space. Using the reparameterization trick \cite{kingma2022autoencodingvariationalbayes}, we sample the latent variable and project it to the embedding space of the CFM model using a single trainable MLP layer. These feature embeddings are added (Fig.~\ref{fig:schematic}) to the time embeddings and passed to the U-Net. We use the same U-Net architecture as the ICFM for all experiments. The model is trained using the loss in Eq.~\ref{eq:auxvae_mainloss} to enforce disentanglement in the latent spaces. Other hyperparameters and their fixed values are presented in Table~\ref{tab:hyperparams}. For both models, inference was performed using the adaptive \texttt{dopri5} solver.

The code for training and evaluation of \texttt{DL-CFM} can be found in \url{https://anonymous.4open.science/r/Latent_CFM-66CF}.

\begin{table}
\center
\begin{tabular}{|c|c|}
\hline
\textbf{Hyperparameters} & \textbf{tSZ data} \\
\hline
Train set size & 10,142  \\

\# steps & 240K  \\

Training batch size & 128   \\

Optimizer & Adam   \\

Learning rate & 2e-4  \\

Latent dimension & 256 \\

Number of model channels & 128 \\

Number of residual blocks & 2 \\

Channel multiplier &  [1, 2, 2, 2] \\

Number of attention heads & 4 \\

Dropout & 0.1 \\

($\beta, \lambda_1, \lambda_2$) & (8e-5, 8e-2, 1e-2)\\

Probability path $\sigma$ & 0 \\
\hline
\end{tabular}
\vspace{0.2cm}
\caption{Hyperparameter settings used for \texttt{DL-CFM} model training on the tSZ dataset.}
\label{tab:hyperparams}
\end{table}

\subsection{Computational cost}
Both ICFM and \texttt{DL-CFM} models were trained using NVIDIA A100 GPUs. For both models, it took $\sim 24$ hours to complete $240$K training steps.

\section{Additional experimental results} \label{sec:additional_results}
\begin{figure}
    \centering
    \includegraphics[width=0.95\linewidth]{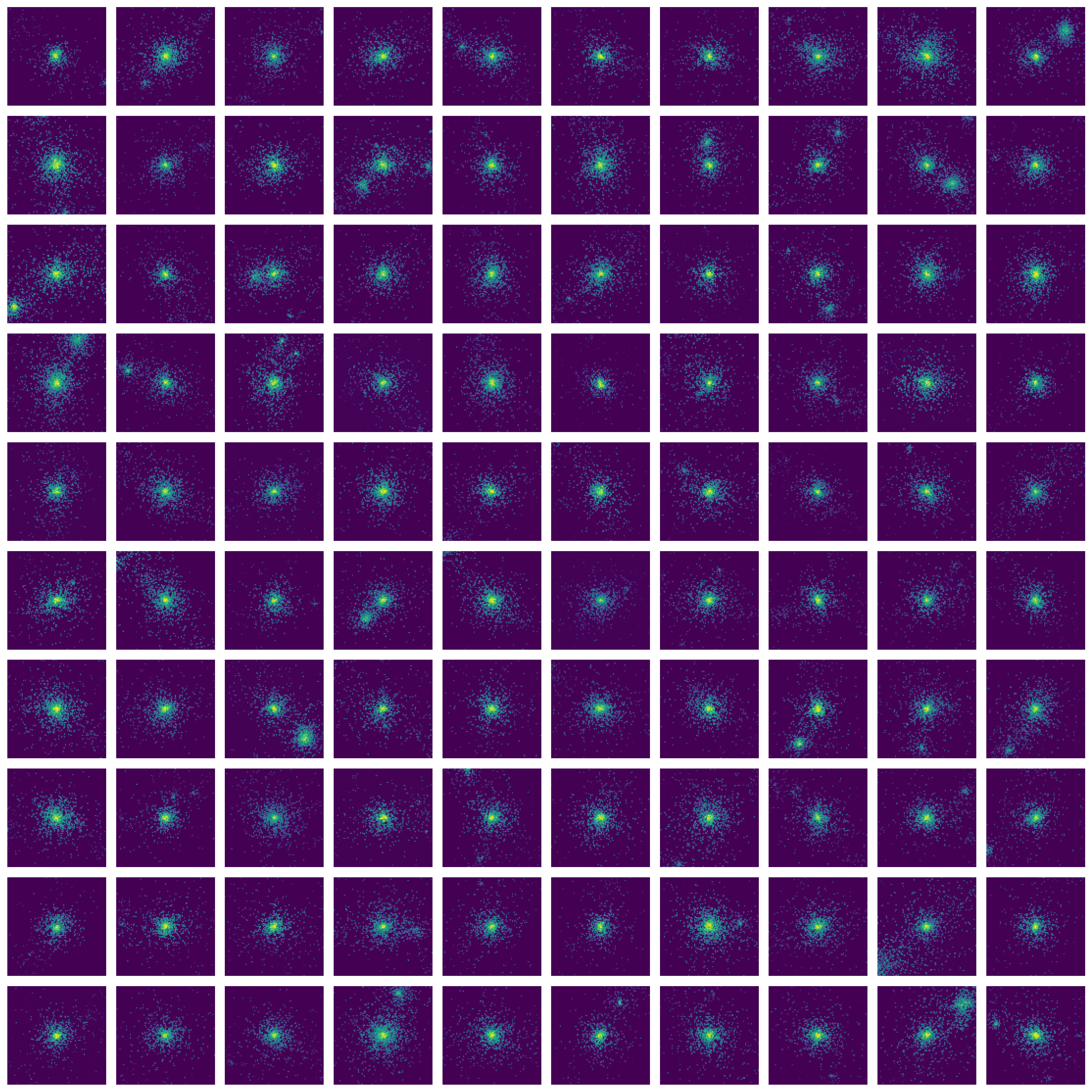}
    \caption{Generating samples from the \textit{center} of the reconstruction-focused latents $z_{\mathrm{rec}}$, with the first two auxiliary-guided coordinates fixed at $(z_1,z_2)=(0.001,0.9)$ - \textbf{low-mass high-concentration setting}.}
    \label{fig:placeholder}
\end{figure}

\begin{figure}
    \centering
    \includegraphics[width=0.95\linewidth]{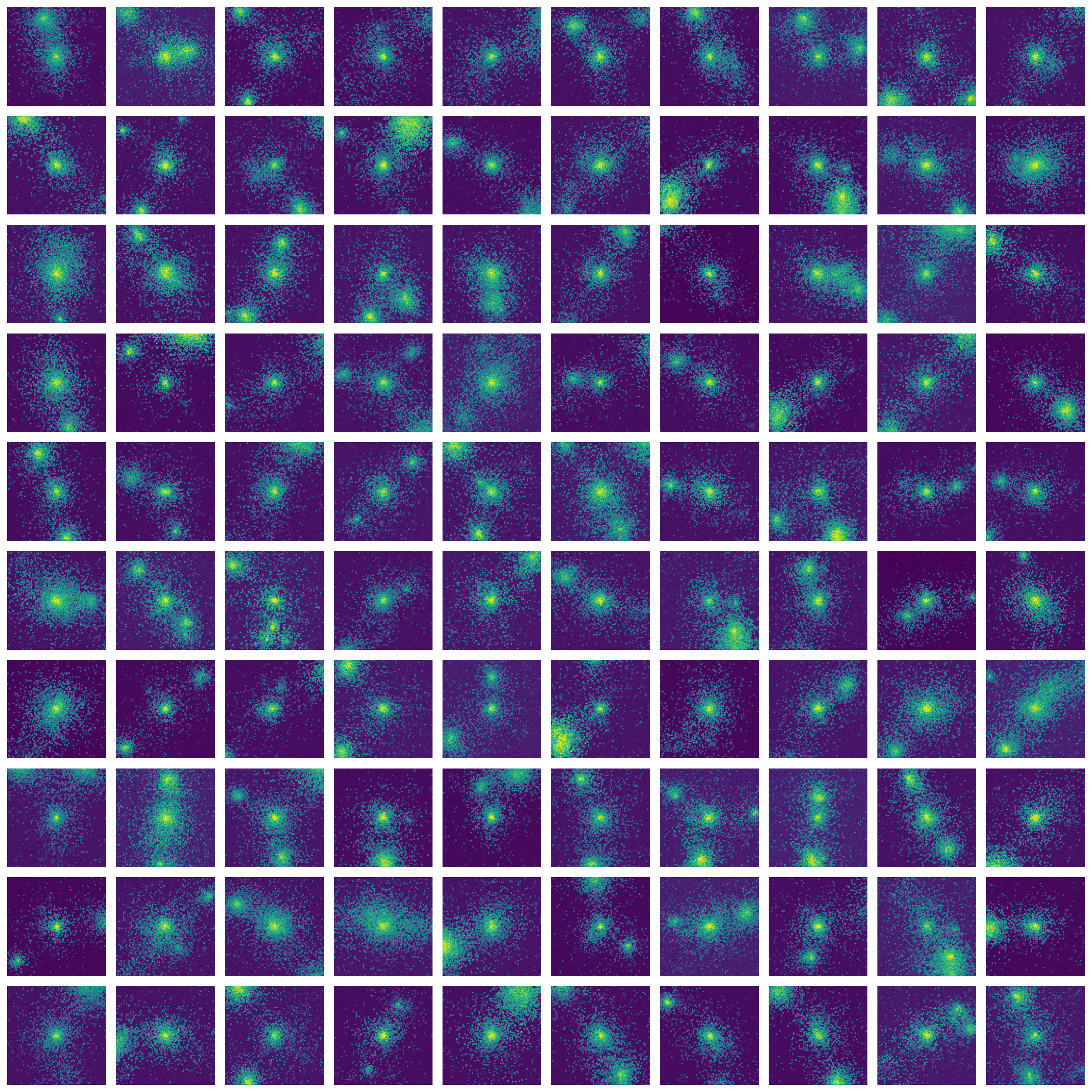}
    \caption{Generating samples from the \textit{tail} of the reconstruction-focused latents $z_{\mathrm{rec}}$, with the first two auxiliary-guided coordinates fixed at $(z_1,z_2)=(0.001,0.9)$ - \textbf{low-mass high-concentration setting}.}
    \label{fig:placeholder}
\end{figure}

\begin{figure}
    \centering
    \includegraphics[width=0.95\linewidth]{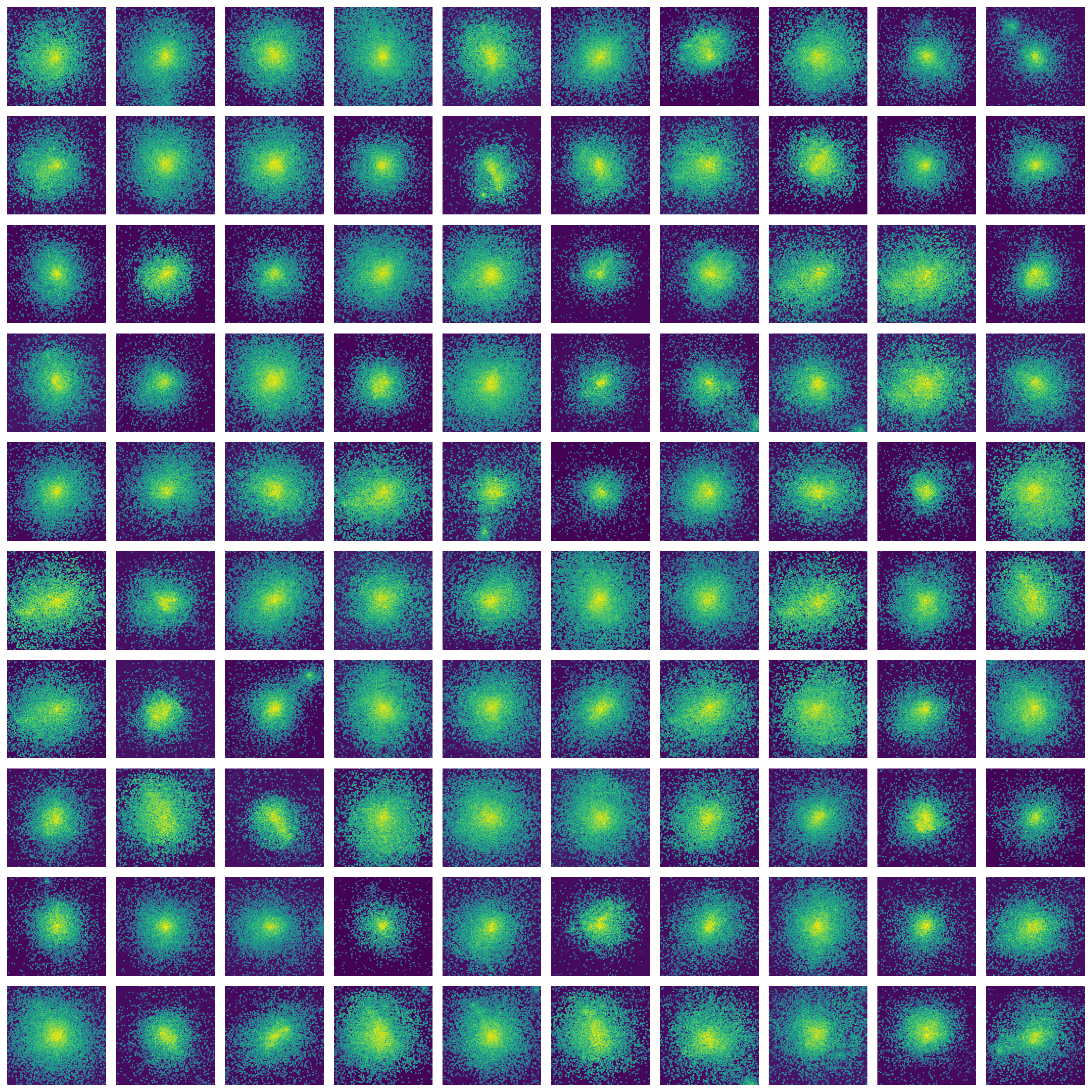}
    \caption{Generating samples from the \textit{center} of the reconstruction-focused latents $z_{\mathrm{rec}}$, with the first two auxiliary-guided coordinates fixed at $(z_1,z_2)=(0.9,0.001)$ - \textbf{high-mass low-concentration setting}.}
    \label{fig:placeholder}
\end{figure}

\begin{figure}
    \centering
    \includegraphics[width=0.95\linewidth]{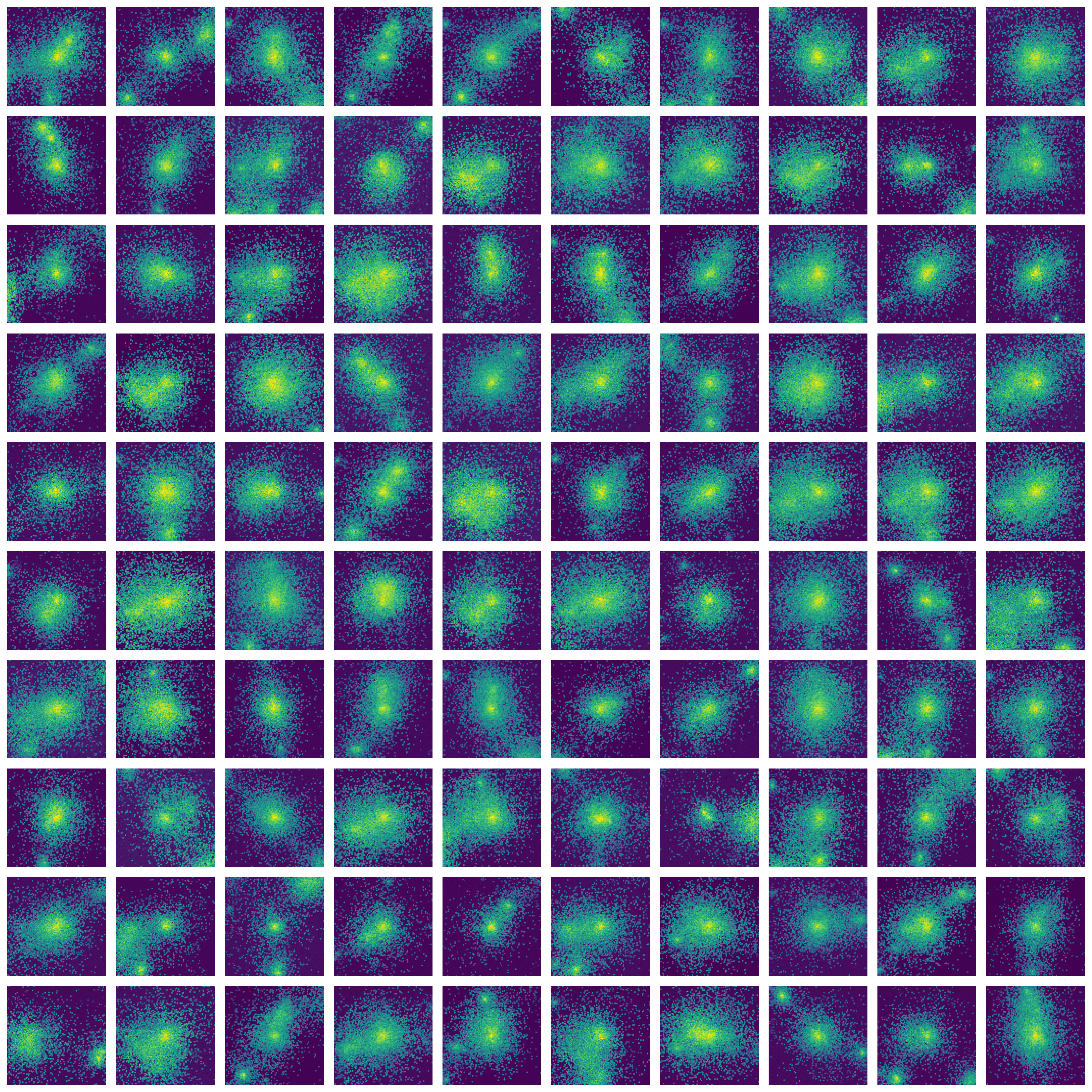}
    \caption{Generating samples from the \textit{tail} of the reconstruction-focused latents $z_{\mathrm{rec}}$, with the first two auxiliary-guided coordinates fixed at $(z_1,z_2)=(0.9,0.001)$ - \textbf{high-mass low-concentration setting}.}
    \label{fig:placeholder}
\end{figure}

\begin{figure}
    \centering
    \includegraphics[width=0.5\linewidth]{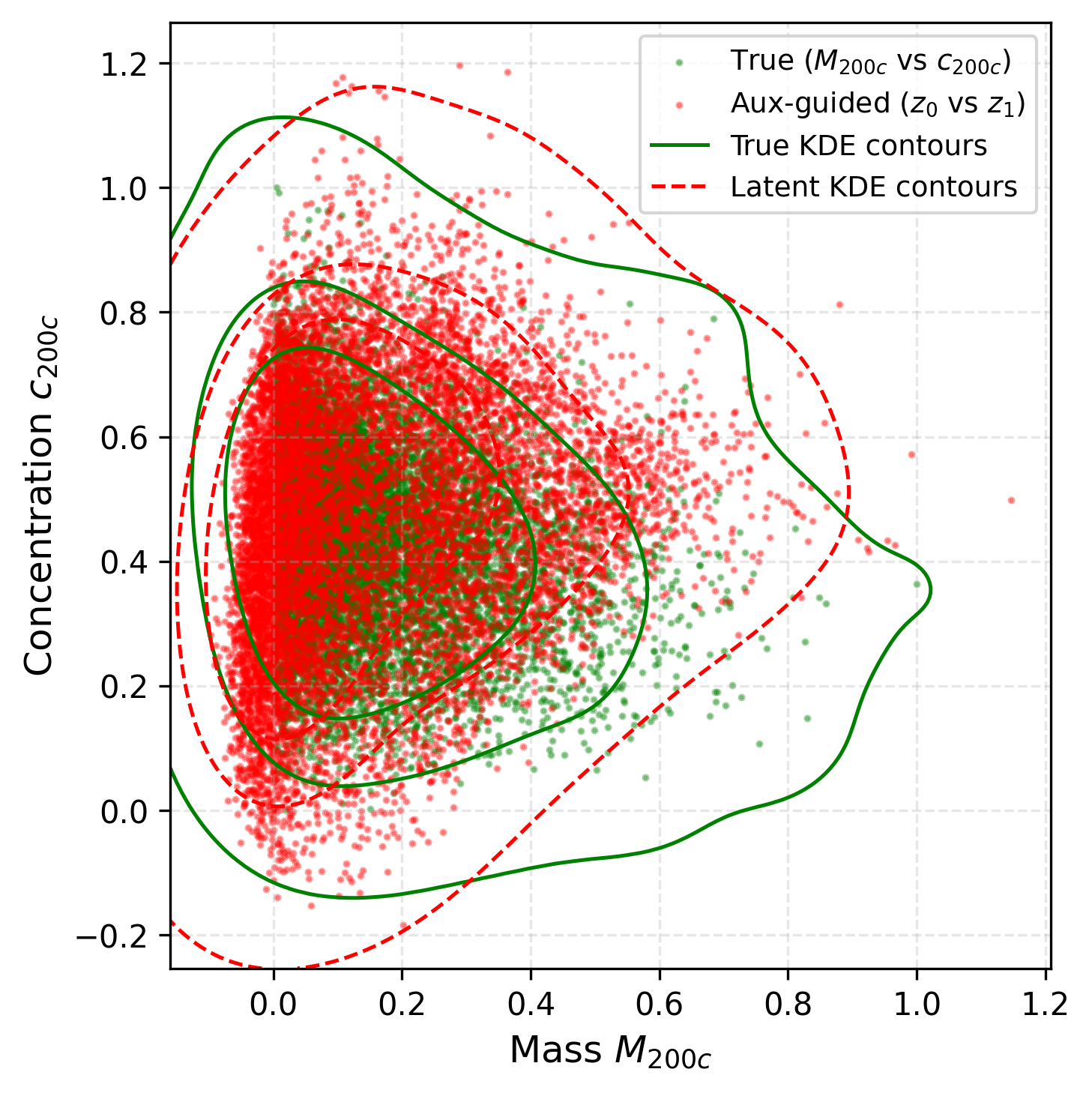}
    \caption{The auxiliary-guided latents maintain the inter-dependency between halo mass and concentration, yielding consistent $M_{200c}$–$c_{200c}$ dependency structure.}
    \label{fig:z1_z2_scatter}
\end{figure}

\end{document}